\documentclass[final,12pt, a4paper]{elsarticle}

\usepackage{enumitem}
\usepackage{amsmath,amssymb,amsfonts}
\usepackage{subfigure}
\usepackage{algorithmic}
\usepackage{graphicx}
\usepackage{textcomp}
\usepackage{xcolor}
\usepackage{tikz}
\usepackage{hyperref}
\usepackage[flushleft]{threeparttable}
\usepackage{array,booktabs,makecell}
\usepackage{listings}
\usepackage{amssymb}
\usepackage{xcolor} 
\usepackage{tcolorbox} 
\usepackage{caption}
\usepackage{tcolorbox} 

\usepackage{titlesec} 
\usepackage{hyperref} 
\usepackage{multicol} 
\usepackage{graphicx} 
\usepackage{array}    
\usepackage{booktabs} 
\usepackage{pifont}   

\usepackage{newunicodechar}

\newcommand{\cmark}{\ding{51}}  
\newcommand{\xmark}{\ding{55}}  

\journal{SoftwareX}
\begin{document}
\renewcommand{\labelenumii}{\arabic{enumi}.\arabic{enumii}}

\begin{frontmatter}



\title{DREAMS: {A python framework for Training Deep Learning Models on EEG Data  with Model Card Reporting for Medical Applications}}


\author[label1]{Rabindra Khadka}
\author[label1,label2]{Pedro G.~Lind}
\author[label1]{Anis Yazidi}
\author[label1]{Asma Belhadi}
\address[label1]{Department of Computer Science, OsloMet -- Oslo Metropolitan University, N-0130 Oslo, Norway}
\address[label2]{Simula Research Laboratory, Numerical Analysis and Scientific Computing, N-0164 Oslo, Norway}

\begin{abstract}

Electroencephalography (EEG) provides a non-invasive way to observe brain activity in real time. Deep learning has enhanced EEG analysis, enabling meaningful pattern detection for clinical and research purposes. However, most existing frameworks for EEG data analysis are either focused on preprocessing techniques or deep learning model development, often overlooking the crucial need for structured documentation and model interpretability.
In this paper, we introduce DREAMS (Deep REport for AI ModelS), a Python-based framework designed to generate automated model cards for deep learning models applied to EEG data. Unlike generic model reporting tools, DREAMS is specifically tailored for EEG-based deep learning applications, incorporating domain-specific metadata, preprocessing details, performance metrics, and uncertainty quantification. The framework seamlessly integrates with deep learning pipelines, providing structured YAML-based documentation. We evaluate DREAMS through two case studies: an EEG emotion classification task using the FACED dataset and a abnormal EEG classification task using the Temple Univeristy Hospital (TUH) Abnormal dataset. These evaluations demonstrate how the generated model card enhances transparency by documenting model performance, dataset biases, and interpretability limitations. Unlike existing model documentation approaches, DREAMS provides visualized performance metrics, dataset alignment details, and model uncertainty estimations, making it a valuable tool for researchers and clinicians working with EEG-based AI. The source code for DREAMS is open-source, facilitating broad adoption in healthcare AI, research, and ethical AI development.


\end{abstract}


\end{frontmatter}


\section{Introduction}

Electroencephalography (EEG)-based deep learning has demonstrated significant potential in various applications, including brain-computer interfaces (BCI), emotion recognition, and neurological disorder diagnosis \cite{hosseini2020review,aminoff2012electroencephalography,islam2023recent}. Despite these advancements, a key challenge in the development and deployment of deep learning models for EEG applications is the lack of structured and standardized model documentation. Models designed for EEG based applications are highly sensitive to variations in dataset preprocessing, feature extraction techniques, and model architectures, making comparability and reproducibility challenging~\cite{mcdermott2021reproducibility}.

Reproducibility is a fundamental requirement in healthcare applications of machine learning (ML) models, yet  EEG-based deep learning research and applications continues to suffer from inconsistent and non-transparent model reporting practices. Many studies report only high-level performance metrics (e.g., accuracy, F1-score) while omitting crucial details such as preprocessing pipelines, hyperparameter configurations, and prediction uncertainties. The absence of structured documentation not only hinders reproducibility but also complicates model benchmarking and comparative analysis, limiting the ability of researchers to validate findings across different datasets and experimental settings. This challenge becomes even more critical in clinical applications, where a lack of reproducibility can impede regulatory approval, deployment, and real-world patient outcomes. Addressing these gaps requires a standardized and automated model documentation framework that ensures transparent, structured, and reproducible reporting of EEG-based deep learning models.

One of the key challenges in deploying deep learning models is their inherent uncertainty in predictions, often leading to overconfidence in incorrect classifications \cite{kendall2017uncertainties,sensoy2018evidential}. This issue is particularly critical in EEG-based deep learning, where high signal variability, inter-subject differences, and noise hinder model generalization capacity \cite{xu2020cross}. In clinical settings, failing to quantify uncertainty can result in misdiagnoses or incorrect treatments, posing high risks \cite{mcdermott2021reproducibility}.

Uncertainty estimation enhances risk-aware decision-making, allowing models to flag low-confidence predictions for further review instead of making potentially harmful automatic decisions. For example, an EEG-based seizure detection system with uncertainty awareness can alert clinicians when a prediction is inconclusive, prompting manual verification. Integrating uncertainty estimation into structured model documentation, such as model cards, ensures greater transparency, interpretability, and trustworthiness in EEG-based healthcare AI \cite{mitchell2019model}.
\begin{figure*}[t] 
\centering
\includegraphics[width=\textwidth, height=0.5\textheight, keepaspectratio]{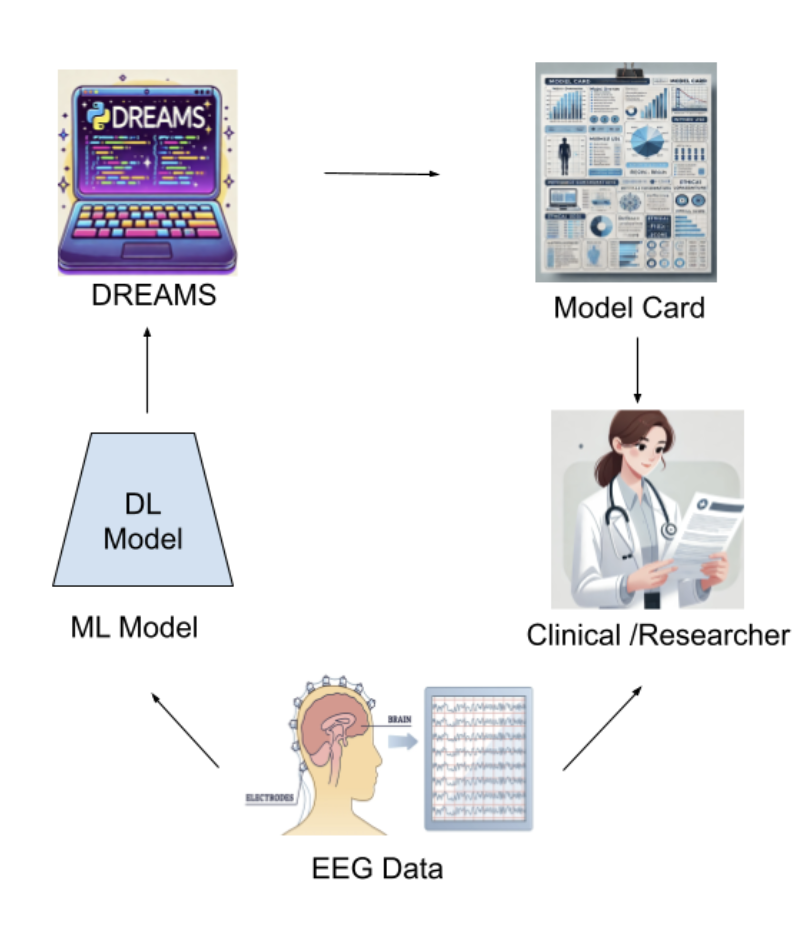}
\caption{\protect 
   Overview of the potential role and usefulness of DREAMS for EEG data analysis in the context of clinical research.} \label{fig:schema}
\end{figure*}

 To address these challenges, we introduce DREAMS, a Python-based model card framework specifically designed for EEG-based deep learning models. DREAMS provides automated documentation covering key aspects such as dataset characteristics, preprocessing techniques, model architecture, performance evaluation with confidence intervals, and version tracking. 

 Our main contributions are:
\begin{enumerate}
    \item We propose DREAMS, a structured model card framework for EEG deep learning, enabling comprehensive documentation of model performance, preprocessing pipelines, and uncertainty estimation.
 \item We introduce confidence interval-based uncertainty reporting, improving model interpretability and helping researchers assess the prediction reliability of the models.
  \item We automate model documentation with version tracking, allowing for efficient benchmarking and reproducibility across different model versions and EEG datasets.
 \item We demonstrate the framework's applicability in EEG-based classification tasks, showcasing its potential for broader EEG-based ML applications.

 \end{enumerate}

\begin{table*}[t]
\centering 
\footnotesize
\begin{tabular}{p{0.3\linewidth}p{0.4\linewidth}}
  \toprule \toprule
  Current code version & v0.0.3  \\
  \midrule
  Permanent link to code/repository used for this code version & 
  \url{https://github.com/LucidJun/DREAM} \\
  \midrule 
  \midrule 
  Legal Code License   & Released under BSD 3-Clause License \\
  \midrule 
  Code versioning system used &  git \\
  \midrule 
  Software code languages, tools, and services used & Python, Java, CSS, SCSS \\
 \midrule  
 If available Link to developer documentation/manual &  \url{https://dreams-mc.readthedocs.io/en/latest/index.html} \\
 \midrule 
 Support email for questions & rabindra@oslomet.no \\
 \bottomrule\bottomrule
 \end{tabular}
\caption{\protect Code metadata of the Python library DREAMS.}
 \label{codeMetadata} 
\end{table*}

Figure \ref{fig:schema} sketches DREAMS role in the context of clinical research based in EEG data.
Table \ref{codeMetadata} provides the main metadata of the code of DREAMS.

\section{Related work}

Some previous works must be highlighted, which will better contextualize and motivate the novelty of DREMS. 
Moons et al. \cite{moons2015transparent} introduced the Transparent Reporting of Multivariable Prediction Model for Individual Prognosis or Diagnosis (TRIPOD) statement, which aimed to enhance the reporting of prediction models. The TRIPOD statement is a checklist comprising items essential for the transparent reporting of prediction model studies. This explanation document details the rationale behind each item, clarifies their meanings, and serves as a valuable reference for considerations when reporting the design and analysis of prediction model studies.

Sendak et al. \cite{sendakpresenting} introduced the Model Facts label, designed for clinicians who make decisions supported by machine learning models. Its purpose is to collect relevant, actionable information into a single page. Its major sections include the model name, locale and version, a summary of the model, the mechanism of risk score calculation, validation and performance metrics, usage directions, warnings and additional information. 

Crisan et al. \cite{crisan2022interactive}  explored interactive model cards, which enhance traditional static model cards by allowing users to explore documentation and interact with models. It includes a conceptual study with ML, NLP and AI Ethics experts, followed by an evaluative study with non-expert analysts.

Mitchell et al. \cite{mitchell2019model} presented a framework for transparent model reporting, primarily targeting human-centered models in computer vision and natural language processing, though it can be applied to any machine learning model. They illustrated the concept using model cards for two examples: one that detects smiling faces in images and another that identifies toxic comments in text. All of the architectures mentioned have certain limitations: they may oversimplify complex details and are often tailored to specific settings. Some of these approaches require technical infrastructure for implementation and others are presented as guidelines on particular domains, limiting their broader applicability.

\subsection{Comparison with Existing Model Documentation Frameworks}

The increasing adoption of deep learning in EEG analysis necessitates structured documentation tools that ensure transparency, reproducibility, and ethical compliance. While existing model documentation frameworks offer general-purpose solutions, they often lack domain-specific adaptability for EEG-based AI. This section compares DREAMS with prominent model documentation approaches, highlighting its unique contributions.

Several model documentation framework exists, but they are primarily designed for general machine learning models rather than the specialized needs of EEG-based deep learning. One such framework is the Google’s Model Card Toolkit (MCT)~\cite{mct} that provides structured model reporting, but it is heavily integrated with TensorFlow Extended (TFX) and lacks direct support for EEG preprocessing metadata or domain-specific performance metrics. Similarly,  Hugging Face Model Cards~\cite{hf} have become a standard for documenting NLP and computer vision models, but they do not accommodate the interpretability challenges, uncertainty quantification, or ethical considerations unique to EEG-based deep learning.
While the traditional manual model documentation methods~\cite{moons2015transparent} (e.g., static reports, research papers) is flexible but lacks automation, standardization, and reproducibility, making it time-consuming and inconsistent across studies.
Without a structured and automated approach, documenting EEG-based deep learning models remains inefficient and prone to inconsistencies, underscoring the need for a dedicated framework like DREAMS.

Unlike these frameworks, DREAMS is specifically designed for EEG-based deep learning models and offers several key advantages, namely:
\begin{itemize}
\item \textbf{EEG-Specific Adaptation}. DREAMS includes EEG preprocessing details, dataset alignment reports, and uncertainty estimation, ensuring domain-relevant transparency.
\item\textbf{Automated Documentation}. Unlike manual reporting, DREAMS automatically logs model parameters, training details, and performance metrics, reducing human error and improving reproducibility.
\item\textbf{Framework Agnostic}. Unlike Google’s Model Card Toolkit, which is tied to TensorFlow, DREAMS works with any deep learning framework (e.g., PyTorch, Keras, TensorFlow).
\item\textbf{Enhanced Interpretability}. DREAMS provides visualized model performance metrics, confidence intervals, and dataset biases, making it more informative than text-heavy model cards.
\item\textbf{Ethical and Regulatory Compliance}. By documenting bias risks, dataset composition, and interpretability challenges, DREAMS aligns with best practices in responsible AI for medical applications.
\end{itemize}

Table~\ref{tab:model_compare} summarizes the differences between DREAMS and existing model documentation approaches.
\begin{table}[!htb]
    \centering
    \renewcommand{\arraystretch}{1.3}
    \resizebox{\textwidth}{!}{%
    \begin{tabular}{|p{3.5cm}|p{2cm}|p{2cm}|p{2cm}|p{2cm}|}
        \hline
        \textbf{Feature} & \textbf{Google MCT} & \textbf{Hugging Face} & \textbf{Manual Docs} & \textbf{DREAMS} \\
        \hline
        Designed for EEG Models? & \xmark{} (No) & \xmark{} (No) & {} (Limited) & \cmark{} (Yes) \\
        \hline
        Automated Model Card Generation & {} (Partial, TFX) & \xmark{} (No) & \xmark{} (No) & \cmark{} (Yes) \\
        \hline
        Integration with DL Frameworks & TensorFlow Only & PyTorch /TensorFlow & \xmark{} (No) & Works with Any DL Framework \\
        \hline
        Dataset Transparency & Basic Metadata & Basic
        Metadata & {} (Manual) & EEG-Specific Metadata \\
        \hline
        Performance Metrics Logging & {} (Limited) & \cmark{} (Yes) (for NLP) & \xmark{} (No) & \cmark{} (Yes, Integrated) \\
        \hline
        Uncertainty Estimation & \xmark{} (No) & \xmark{} (No) & \xmark{} (No) & \cmark{} (Yes) (Confidence Intervals) \\
        \hline
        Ethical Considerations & \cmark{} (Encouraged) &  \cmark{} (Encouraged) & Researcher-Dependent & \cmark{} (EEG Bias Reporting) \\
        \hline
        Visualization of Results & \xmark{} (No) & \xmark{} (No) & \xmark{} (No) & \cmark{} (Performance Plots) \\
        \hline
        Reproducibility & {} (Limited, TFX Required) & \xmark{} (No Tracking) & \xmark{} (No Tracking) & \cmark{} (YAML-Based Logging) \\
        \hline
        Customization & {} (Limited) & {} (Limited to NLP) & \xmark{} (No Format) & \cmark{} (Flexible YAML) \\
        \hline
        Open Source? & \cmark{} (Yes) & \cmark{} (Yes) & \xmark{} (No) & \cmark{} (Yes) \\
        \hline
    \end{tabular}
    }
    \caption{Comparison of Model Documentation Frameworks}
    \label{tab:model_compare}
\end{table}

\section{Software description}

To illustrate the concept of DREAMS, we present a case study in which model cards (MC) are utilized to highlight the key aspects of the data, model, and performance metrics, as well as the limitations encountered in the study. 
MCs serve as a detailed documentation tool, emphasizing the characteristics and context of the data used, the specific features and architecture of the model and reports its performance. Moreover, it can include information about limitations or challenges identified during the research. The use of MCs when developing DL underscores the importance of transparency and accountability in model development and deployment.

\subsection{Software architecture}
DREAMS, developed as a Python package, offers a comprehensive
framework aimed at simplifying the process of creating comprehensive model cards for deep learning models. 
Model cards (MC) serve as vital documentation tools, outlining the performance metrics, usage guidelines and limitations of AI models. They play a pivotal role in enhancing transparency and promoting the ethical utilization of AI technologies.

\begin{figure*}[t] 
\centering
\includegraphics[width=\textwidth]{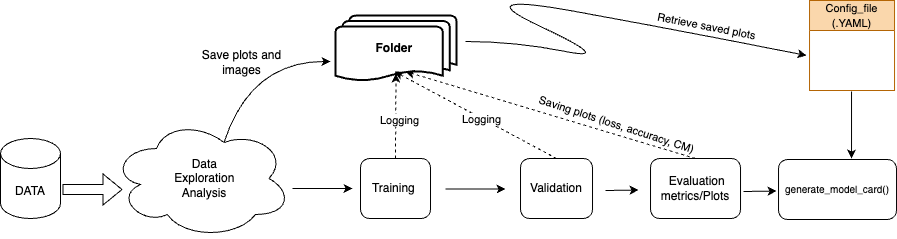}
\caption{
  \textbf{Workflow of the DREAMS Model Card Generation Process}: The diagram illustrates the end-to-end pipeline, starting from data collection and exploration, followed by training and validation phases, and progressing to evaluation metrics and plot generation. Key outputs, including plots and logs, are saved in a designated folder, which are then retrieved by the configuration file (.YAML). Finally, the \texttt{$generate\_model\_card()$} function utilizes the configuration and saved assets to automatically generate the structured model card.}  \label{fig:pipeline}
\end{figure*}

The DREAMS architecture is modular, comprising several key components. It starts with exploratory data analysis (EDA) on the provided dataset, with the visualizations saved in a designated folder. As the model progresses through training and validation, this folder is further enriched with additional plots, such as loss, accuracy and confusion matrix. Finally, the model card function is invoked, using a configuration file that contains all the necessary details to generate the model card. An overview of the architecture of DREAMS' library is shown in Figure \ref{fig:pipeline}.

\subsection{Software functionalities}

\subsubsection{Data preprocessing }

The pipeline typically begins with data collection and preprocessing, where the raw data is gathered from various sources and prepared for model input. 
Preprocessing steps may include normalization, data augmentation and transformation into formats that are compatible with the model, such as converting images to tensors or encoding text data.
Then, the data is usually divided into training and validation sets. The training set is used to train the model, while the validation set assists fine-tune the hyperparameters of the model.
To ensure that every step in this process is well-documented and reproducible, we systematically store all critical details about the data, including information on preprocessing, in a designated folder. 
In a YAML file, we define the exact path to this folder, ensuring that the information is easily accessible and organized for future reference or use. 
This DREAMS framework setup not only facilitates transparency but also enables efficient tracking of data-related processes throughout the entire model development pipeline.

\subsubsection{Model training and evaluation} 

The training pipeline of a deep learning model is a comprehensive process that transforms raw data into a well-trained model capable of making accurate predictions or classifications.
During the training phase, the model processes batches of data iteratively, adjusting its internal parameters to minimize the error between its predictions and the actual outcomes. This is achieved through an optimization algorithm like stochastic gradient descent, which updates the model's parameters based on the gradients of the loss function.
Throughout the training pipeline, performance metrics such as accuracy, loss, precision, recall, and F1-score are monitored to assess the model's progress. Visualization tools such as confusion matrices are often used to gain deeper insights into the model’s performance.
The model is then ready for inference, where it can be used to make predictions on new, unseen data. 

At this stage, we meticulously document all the paths to vital information related to both training and evaluation results within the YAML file. 
This DREAMS comprehensive record encompasses not only paths to graphics but also to other essential data, ensuring that every critical component is easily accessible and well-organized. This approach ensures efficient management and retrieval of information throughout the training and evaluation processes.

\subsubsection{Model card generation}

Once all the necessary information for generating the model card has been gathered, we proceed to invoke the \textit{generate\_modelcard()} function. 
This function requires three input parameters: the path to the YAML configuration file, which contains the essential details for the model card; the output path, where the generated model card will be saved; and the version number of the model card, which helps track its iterations and updates.

An example of the code snippet  using \textit{generate\_modelcard()} function is as follows: 
\begin{lstlisting}
# model_card library
from dreams_mc.make_model_card import generate_modelcard 

config_file_path="./config.yaml"
output_path= "./logs/model_card.html"
version_num = '1.0'

# Generate the model card
generate_modelcard(config_file_path,output_path,version_num)
\end{lstlisting}

\section{Hypothesis and Limitations}

The primary hypothesis of this work is that structured documentation of deep learning models through \textit{automated model cards} can significantly enhance \textit{transparency, reproducibility, and ethical compliance} in EEG-based deep learning. Specifically, we hypothesize that:
\begin{itemize}
    \item A domain-specific model card frameworks (DREAMS) tailored for EEG-based deep learning can improve the reporting of performance metrics, dataset biases, and model uncertainty compared to generic model documentation approaches.
    \item Automated documentation of EEG preprocessing steps, dataset details, and evaluation results will reduce inconsistencies and improve reproducibility in deep learning-based EEG research.
    \item A standardized model card structure incorporating EEG-specific metadata can facilitate better interpretability and regulatory compliance in healthcare AI applications.
\end{itemize}

To support this hypothesis, we make some assumptions regarding the use of EEG data and deep learning models:
\begin{itemize}
    \item The quality of model documentation is a critical factor affecting the interpretability and trustworthiness of EEG-based deep learning models.
    \item The datasets used in this study (FACED and CAUEEG) are representative of real-world EEG classification tasks, allowing generalization of results to similar applications.
    \item The reported performance metrics (e.g., accuracy, F1-score, uncertainty estimates) provide meaningful insights into model effectiveness and limitations, which are accurately captured in the generated model cards.
    \item Researchers and clinicians will benefit from structured model documentation without significant modifications to existing deep learning workflows.
\end{itemize}

\subsection{Limitations}
Despite the advantages of DREAMS, we acknowledge some limitations. The framework's effectiveness depends on the EEG \textit{data quality} and preprocessing techniques. Poorly preprocessed or noisy EEG signals may introduce biases that cannot be fully mitigated through documentation alone. Additionally, while DREAMS automates model documentation, the logging and visualization processes introduce \textit{computational overhead}, which may not be ideal for real-time EEG applications requiring low-latency performance. Another limitation is the framework’s current level of \textit{clinical validation}; while it is designed for research and development, its direct applicability in clinical environments requires further evaluation with real-world patient data and potential regulatory compliance assessments. Furthermore, although DREAMS is framework-agnostic, its implementation is \textit{primarily optimized} for Python-based deep learning libraries such as PyTorch and TensorFlow, limiting its immediate usability for researchers working with other AI ecosystems. Lastly, while DREAMS promotes transparency by documenting model biases and performance limitations, it does not eliminate the \textit{ethical concerns} associated with EEG-based deep learning models. Ensuring fair data collection, privacy protection, and algorithmic fairness remains the responsibility of researchers and developers implementing the framework. By recognizing these challenges, we highlight areas for future improvements and refinements, ensuring DREAMS continues to evolve as a robust and practical model documentation tool for EEG-based AI applications.

\vspace{2em}
\section{Illustrative Examples on EEG data} \label{sec:illustEEG}

We provide examples of model cards for two models: an EEG-based emotion classifier and an EEG dementia classification model.

\subsection{Emotion Classifier}

We present a model card for an EEG-based emotion classifier trained on the Finer-grained Affective Computing EEG Dataset (FACED) \cite{chen2023large}, which is compatible with TorchEEG \footnote{https://torcheeg.readthedocs.io/en/latest/}. The dataset consists of EEG recordings from 123 subjects using 32 electrodes positioned according to the international 10–20 system. During the recording sessions, participants viewed 28 different video clips designed to elicit various emotional responses. These video clips spanned nine distinct emotion categories: amusement, inspiration, joy, tenderness, anger, fear, disgust, sadness, and neutral emotion. We applied band-pass filtering, common spatial pattern filtering, and normalization to the EEG data.

\begin{tcolorbox}[colframe=black, colback=white, sharp corners, boxrule=1pt, width=\textwidth]
\scriptsize 

\begin{center}
    { \textbf{ Model Card - EEG Emotion Classifier (v 0.1.0)}}
\end{center}

\begin{multicols}{2}

\section*{\textcolor{green!50!black}{Overview}}
This model classifies EEG signals into \textbf{negative}, \textbf{neutral}, and \textbf{positive} classes. It predicts emotions such as \textit{anger, disgust, fear, amusement, joy}, and more. Additionally, it provides uncertainty estimation of predictions with confidence intervals, hyperparameters, and performance metrics.

\section*{\textcolor{green!50!black}{Dataset}}
\begin{itemize}[noitemsep,topsep=0pt]
    \item \textbf{Dataset:} FACED Dataset
    \item \textbf{Number of Classes:} 3
    \item \textbf{Ground Truth:} Negative (0), Neutral (1), Positive (2)
    \item \textbf{Training/Validation Split:} 80:20
    \item \textbf{Preprocessing:} Band-Pass Filtering, Common Spatial Patterns (CSP), Normalization
\end{itemize}

\begin{center}
    \includegraphics[width=0.75\columnwidth]{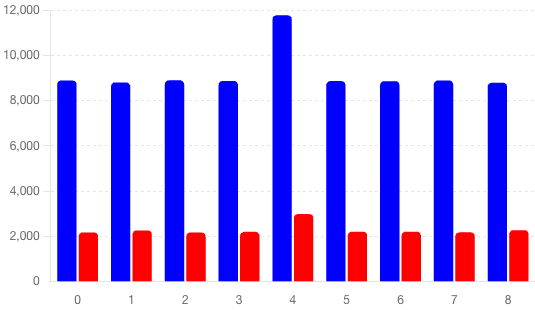} 
\end{center}


\section*{\textcolor{green!50!black}{Model Details}}
\begin{itemize}[noitemsep,topsep=0pt]
    \item \textbf{Input:} 30-channel EEG time-series data
    \item \textbf{Output:} Class labels, Confidence intervals
    \item \textbf{Model Type:} CNN (Xception-based)
    \item \textbf{Learning Rate:} 0.001
    \item \textbf{Batch Size:} 32
    \item \textbf{Parameters:} 0.56M trainable parameters
\end{itemize}


\section*{\textcolor{green!50!black}{Performance}}
\vspace{-1em}
The model’s performance is evaluated on an independent test set. Below is a summary of the performance by confusion matrix.

\begin{center}
    \includegraphics[width=1\columnwidth]{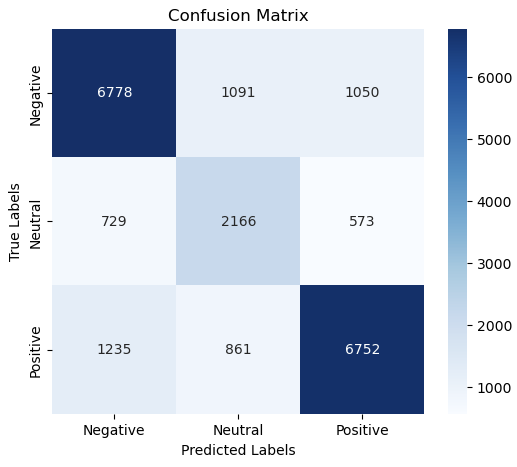} 
\end{center}
\vspace{-3em}

\section*{\textcolor{green!50!black}{Limitations}}
\vspace{-1em}
\begin{itemize}[noitemsep,topsep=0pt]
    \item The dataset exhibits class imbalance (male-to-female ratio 3:1), which might introduce bias.
    \item Noisy EEG channels and missing data can degrade classification performance.
\end{itemize}

\section*{\textcolor{green!50!black}{Uncertainty}}
\vspace{-1em}
Confidence intervals provide insights into model reliability. Below, the performance metrics are reported with 95\% confidence intervals.

\begin{center}
    \includegraphics[width=1\columnwidth]{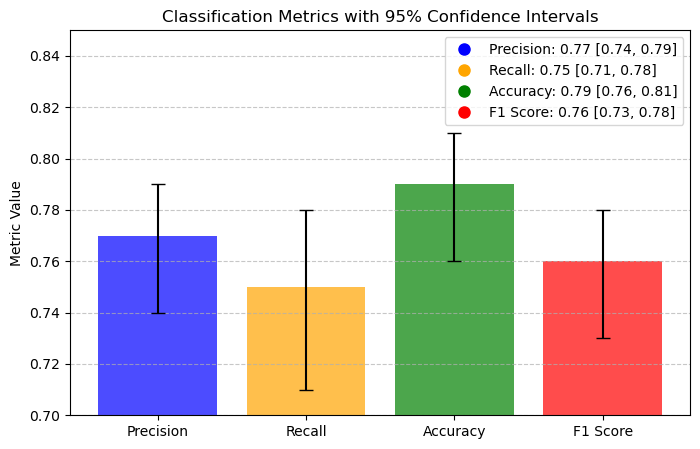} 
\end{center}

\end{multicols} 

\end{tcolorbox} 

\begin{center}\captionof{figure}{\textbf{An illustrative example of Model Card for EEG Emotion Classification.} 
    This model card summarizes the dataset, model details, performance metrics, uncertainity of predictions and limitations for an EEG-based emotion classification system. }
\label{Fig:emot_mc}

\end{center}


\begin{tcolorbox}[colframe=black, colback=white, sharp corners, boxrule=1pt, width=\textwidth]
\scriptsize 

\begin{center}
    { \textbf{Model Card - Abnormal/Normal Classifier (v 1.0.0) }}
\end{center}

\begin{multicols}{2}

\section*{\textcolor{green!50!black}{Overview}}
The model analyzed in this card classifies EEG into abnormal or normal class and returns an uncertainty estimation of the predicted class with a confidence interval. It reports the model's hyperparameters and performance plots. The reader can gain further insights about the model's limitation by identifying the input instances where the model is expected to perform well.

\section*{\textcolor{green!50!black}{Dataset}}
\begin{itemize}[noitemsep,topsep=0pt]
    \item \textbf{Dataset:} TUH Abnormal
    \item \textbf{Number of Classes:} Two
    \item \textbf{Ground Truth:} Health status (abnormal or normal)
    \item \textbf{Training/Validation Split:} 80:20
    \item \textbf{Preprocessing:} Independent Component Analysis, Band-Pass Filtering, Normalization
\end{itemize}

\begin{center}
    \includegraphics[width=\columnwidth]{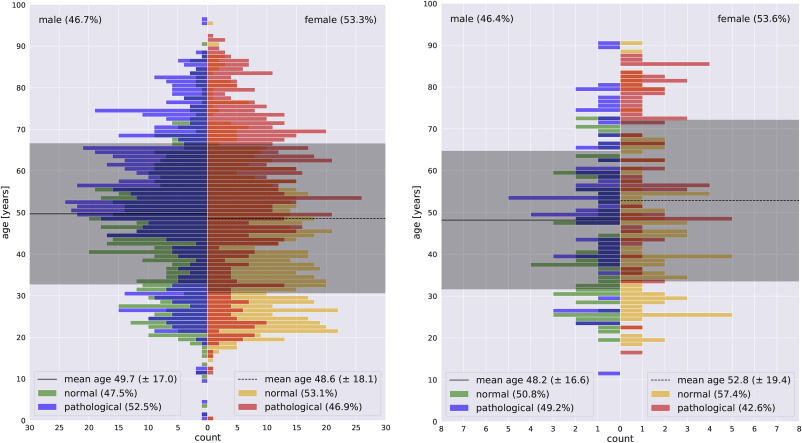} 
\end{center}


\section*{\textcolor{green!50!black}{Model Details}}
\begin{itemize}[noitemsep,topsep=0pt]
    \item \textbf{Input:} 19-channel EEG data in time domain.
    \item \textbf{Output:} Class labels abnormal/normal, Confidence intervals
    \item \textbf{Model Type:} Brain Signal Vision Transformer(BSVT). 
    \item \textbf{Learning Rate:} 1e-05
    \item \textbf{Batch Size:} 32
    \item \textbf{Parameters:} 0.75M trainable parameters
\end{itemize}


\section*{\textcolor{green!50!black}{Performance}}
\vspace{-1em}
The model’s performance is evaluated on an independent test set. Below is  the plot of training and validation accuracy.

\begin{center}
    \includegraphics[width=1\columnwidth]{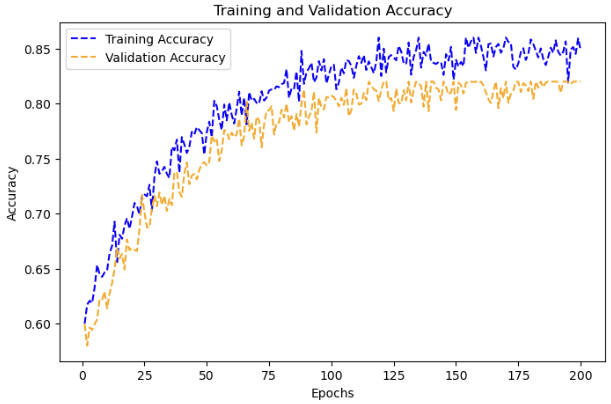} 
\end{center}
\vspace{-3em}

\section*{\textcolor{green!50!black}{Limitations}}
\vspace{-1em}
\begin{itemize}[noitemsep,topsep=0pt]
    \item Long EEG sequences require large tokenized inputs, leading to excessive memory consumption for the transformer model.
    \item Noisy EEG channels and missing data can degrade classification performance.
\end{itemize}

\section*{\textcolor{green!50!black}{Uncertainty}}
\vspace{-1em}
Confidence intervals provide insights into model reliability. Below, the performance metrics are reported with 95\% confidence intervals.

\begin{center}
    \includegraphics[width=1\columnwidth]{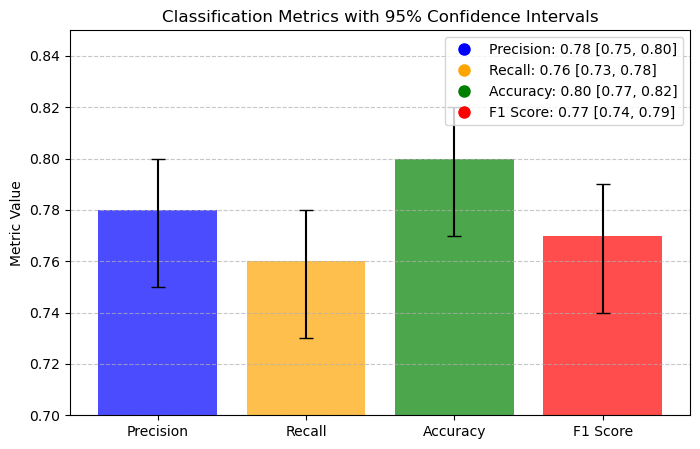} 
\end{center}

\end{multicols} 

\end{tcolorbox} 
\vspace{-1em}
 \begin{center}\captionof{figure}{\textbf{An illustrative example of Model Card for EEG Abnormal/Normal Classification.} 
    This model card summarizes the dataset, model details, performance metrics, uncertainity of predictions and limitations for the classification of abnormal and normal EEG signals. }
    \label{Fig:abnormal_mc}
\end{center}

We applied a deep learning model called TSception, a multi-scale convolutional neural network designed to classify emotions from EEG data. It incorporates dynamic temporal layers, asymmetric spatial layers and high-level fusion layers, enabling the model to learn discriminative representations across both time and channel dimensions simultaneously. The model described in this example classifies EEG data into three main categories: negative (0), which includes emotions such as anger, disgust, fear, and sadness; neutral (1); and positive (2), which includes emotions like amusement, inspiration, joy, and tenderness.

The model card provides a structured evaluation of model performance, including confusion matrix analysis, from which true positive (TP), false positive (FP), true negative (TN), and false negative (FN) rates are derived to assess classification performance. Additionally, it includes uncertainty estimation, offering insights into the confidence levels of model predictions. Furthermore, the model card provides comprehensive documentation, outlining key aspects of the dataset characteristics, preprocessing steps applied to EEG signals, and an overview of the deep learning model for emotion classification.

Figure~\ref{Fig:emot_mc} illustrates the prototype of the model card, highlighting its structured approach to model transparency and interpretability. Notably, the model card is fully customizable and can be extended to incorporate additional domain-specific details, depending on the specific application requirements.

\subsection{Abnormal Classifier}

We provide a second illustrative model card using the publicly available abnormal EEG dataset from the Temple University Hospital (TUH) v2.0.0~\cite{obeid2016temple}. The dataset includes 2,993 recordings from 2,329 different subjects. The recordings are labelled as normal (1,385 recordings) or abnormal (998 recordings). The training set contains 2,717 recordings and the  evaluation set holds 276 recordings. We applied band pass filtering, Independent Component Analysis (ICA), Normalization as the preprocessing steps on the input dataset. 

A transformer-based model, BSVT~\cite{khadka2024inducing}, was employed for the classification task, with its performance and prediction uncertainty systematically documented in the model card. Figure~\ref{Fig:abnormal_mc} presents an example of the model card, showcasing how variations in model architecture and performance evaluation can be structured within a standardized format. Notably, if hyperparameters are modified or an alternative model is selected, the model card can accommodate these updates by specifying the version number and associated changes. Furthermore, the model card is highly customizable, allowing for the inclusion of additional sections as needed. For instance, a recommendation section could be added to provide additional information on model deployment, usage considerations, or the necessity for further validation and testing.

\vspace{1em}

\subsection{Pipeline for Model Card Generation }

This section details the step-by-step process for training a deep learning model using pytorch framework, evaluating its performance, and ultimately generating a structured model card. The pipeline includes data preprocessing, model training, validation, performance analysis, and automated documentation using a model card.

\textbf{Data Preprocessing and Preparation} The EEG data undergoes online transformation into tensor format and a 2D representation for compatibility with deep learning models. The labels are adjusted accordingly, and the dataset is split into training and validation sets. To efficiently manage data batching and shuffling, we create DataLoaders for both training and validation.
The implementation is as follows:

\begin{lstlisting}
data_folder= "./processed_data_Face/Processed_data"

dataset = Dataset(root_path= data_folder,
                   online_transform=transforms.Compose(
                       [transforms.ToTensor(),
                        transforms.To2d()]),
                   label_transform=transforms.Compose([
                       transforms.Select('valence'),
                       transforms.Lambda(lambda x: x + 1)
                   ]))
...

# Dataset Spliting
train_dataset, val_dataset = random_split(dataset, [num_train_samples, num_val_samples])

# DataLoaders creation 
train_loader = DataLoader(train_dataset, batch_size=batch_size, shuffle=True)
val_loader = DataLoader(val_dataset, batch_size=batch_size, shuffle=False)
\end{lstlisting}

\textbf{Model Setup and Configuration} At this stage of the pipeline, the model is initialized and configured with the necessary parameters. The model is instantiated using the \textit{Model function}, specifying key hyperparameters  for instance, in our case the number of classes, electrodes, sampling rate, and architectural components. The following code snippet demonstrates the initialization:

\begin{lstlisting}

model = Model(num_classes=9,num_electrodes=30,
              sampling_rate=250,num_T=15,
              num_S=15, hid_channels=32, dropout=0.5)
\end{lstlisting}

\textbf{Training and Validation Process} We proceed to configure the training and validation processes utilizing PyTorch. This involves setting up the necessary parameters and components for training the model, such as defining the loss functions, optimizers, and evaluation metrics. We establish the procedures for validating the model's performance, including managing the data flow, monitoring metrics, and adjusting the training process as needed to ensure effective learning and evaluation. A standard implement for the training and validation configuration is:
\begin{lstlisting}
# Model Training
valid_acc_max = 0.0
trained_model, accuracy_stats, loss_stats = train(220, valid_acc_max, model, optimizer, criterion, scheduler, train_loader, val_loader, "./logs/current_checkpoint.pt", "./logs/best_model.pt", start_epoch=1)

# plotting and saving training and validation plots
plot_training_validation_stats(accuracy_stats, loss_stats, save_dir='./logs')

# Load the best model for evaluation
checkpoint_path = './logs/best_model.pt'
checkpoint = torch.load(checkpoint_path, map_location=torch.device('cpu'))
model.load_state_dict(checkpoint['state_dict'])
\end{lstlisting}

\textbf{Performance Evaluation and Metrics Visualization} To assess the model’s performance, we generate a confusion matrix and compute key evaluation metrics, such as accuracy, precision, recall, and F1-score. Additionally, uncertainty estimation is applied to quantify the confidence of predictions. 

\begin{lstlisting}
plot_metrics_table(results, model_name='TSception',save_path="./logs/table.png")
plot_confusion_matrix(results['confusion_matrix'], class_names,save_path="./logs/cm.png")
plot_confidence_intervals(precision=results['precision'], recall=results['recall'] ...

\end{lstlisting}

\textbf{Model card Generation} Lastly, we call the \textit{generate\_model\_card()} function, providing it with the path to the YAML configuration file \footnote{https://github.com/LucidJun/DREAMS/tree/main/template}, the output path for saving the model card,  and the version number of the model card as shown below:

\begin{lstlisting}
# Generating model card
print("Generating Model Card ...")
config_file_path = './config.yaml'
output_path = './logs/model_card.html'
version_num = '1.0'
generate_modelcard(config_file_path,output_path,version_num)
\end{lstlisting}

Figure \ref{Fig:emot_mc} and \ref{Fig:abnormal_mc} provide the illustrative example of the generated model cards  following the above describe pipeline. This model card encapsulates the key information and metrics resulting from the model's training and evaluation processes, offering a comprehensive summary of the model's performance, data characteristics, and other relevant details.


\section{Impacts}

In the context of EEG applications, DREAMS offers a smooth solution for generating comprehensive model cards, providing detailed sections that include overviews of EEG analysis projects, dataset descriptions specific to EEG recordings, results with corresponding plots (such as EEG waveforms or brain activity patterns), uncertainty estimations related to signal interpretation, and relevant references. Additionally, users can customize the content of these model cards using a YAML configuration file, enabling adaptability to various EEG-based projects and models, ensuring flexibility in presenting key insights and findings.


DREAMS is entirely model-agnostic, deliberately engineered to be compatible with any type of model. 
Its design ensures versatility, allowing it to generate model cards regardless of the underlying model architecture or type. 
This flexibility is a fundamental aspect of DREAMS, enabling it to adapt to various machine learning and deep learning models, including but not limited to convolutional neural networks (CNNs), recurrent neural networks (RNNs), transformer models and random forests. 
Whether it's a classification, regression or any other type of model, DREAMS generates comprehensive model cards, providing valuable insights and documentation irrespective of the model's complexity or structure. 
The model-agnostic nature of DREAMS enables users to apply it across a diverse array of projects and applications, ensuring its relevance and usefulness in various machine-learning endeavors.

This versatility allows DREAMS to find applicability across a wide range of domains where transparency and ethical use of AI are paramount. 
For instance, within the healthcare sector \cite{djenouri2024artificial,khadka2024inducing}, DREAMS could be utilized to document the performance and limitations of deep learning models used for medical analysis.
Through the generation of detailed model cards, healthcare practitioners and regulatory bodies can gain insights into the model's accuracy, potential biases, and areas of uncertainty, facilitating informed decision-making and ensuring patient safety. 

Similarly, in the financial sector \cite{abakarim2018towards,schumaker2009textual}, DREAMS can aid in documenting AI models used for credit risk assessment or fraud detection, providing transparency into the model's decision-making process and potential biases. 
This transparency is essential for ensuring fairness and accountability in algorithmic decision-making processes.

Furthermore, in intelligent transportation systems \cite{djenouri2023secure}  such as, autonomous vehicles, DREAMS can be employed to document the performance and safety considerations of deep learning models used for object detection and navigation, promoting trust and transparency among regulators and the general public.

DREAMS can be effortlessly installed and utilized via PyPI, ensuring convenient access for users. 
The project's code has been shared on GitHub, allowing for collaborative contributions and enhancements from the community. 
As an open-source initiative, users can extend and customize the software to meet their specific requirements. 
Comprehensive documentation is available in English to assist users in understanding and utilizing the software effectively. 
Note that, the maintenance and upkeep of the software are managed by one of the authors, ensuring ongoing support and reliability.

In essence, DREAMS enables the creation of model cards that hold immense value for AI practitioners. 
These cards provide vital documentation that enhances transparency, facilitates user understanding, and supports responsible deployment by detailing the operational parameters of the model.

\section{Discussions and Future Work}
This work introduces DREAMS, a Python-based model card framework for EEG-based deep learning models, designed to enhance transparency, interpretability, and reproducibility. A key advantage of this framework is the integration of automated model documentation, reducing manual effort while ensuring consistency across model iterations. The incorporation of confidence intervals in model evaluation improves interpretability, enabling researchers to quantify prediction reliability and assess model uncertainty. Furthermore, dataset characteristics, model´s hyperparameters, preprocessing details, and performance analysis can be systematically documented, ensuring comprehensive and standardized reporting across studies.

Beyond improving reproducibility, DREAMS also addresses critical ethical concerns in EEG-based AI by promoting responsible model documentation. Ethical challenges in deep learning, particularly in EEG-based healthcare and neuroscience applications, include bias in training data, lack of transparency in decision-making, and potential risks in clinical deployment. By explicitly documenting dataset demographics, potential biases, and model limitations, the model card enables researchers to assess fairness and mitigate algorithmic bias. Additionally, DREAMS allows for the inclusion of ethical guidelines, ensuring that researchers disclose data handling practices, consent considerations, and limitations of model generalizability, fostering greater accountability in AI-driven EEG applications.

The customizability of DREAMS ensures that model cards can be extended based on specific use cases, incorporating additional sections such as deployment considerations, ethical guidelines, and regulatory compliance requirements. This flexibility allows researchers to tailor documentation for clinical validation, healthcare deployment, or domain-specific performance evaluation, supporting ethical AI development in high-stakes applications.

While model documentation is a significant step toward improving reproducibility, broader challenges remain in EEG-based deep learning. The variability of EEG signals across individuals introduces inherent uncertainty, making generalization difficult. This necessitates future work on subject-adaptive models and personalized EEG feature extraction. Additionally, the lack of standardized reporting practices in EEG-based deep learning further underscores the need for structured model cards as part of a transparent and ethical AI workflow. By systematically addressing uncertainty, bias, and documentation gaps, DREAMS provides a foundational framework for responsible EEG-based AI research and deployment.

Future research should explore the integration of explainability techniques, such as saliency maps and attention mechanisms, to provide deeper insights into EEG feature learning. Furthermore, ensuring fairness and bias detection in EEG models, particularly across different demographics, remains an important direction for future study.

In conclusion, this work provides a structured and automated model card framework designed to improve reproducibility, interpretability, and uncertainty-aware evaluation in EEG-based deep learning. We show the integration of DREAMS into deep learning workflow to enhance model transparency by systematically reporting performance metrics such as accuracy, precision, recall, and F1-score with 95\% confidence intervals. DREAMS ensures automated documentation with version tracking, reducing manual effort and improving reproducibility across model iterations. By addressing the lack of standardized model documentation, our framework ensures that performance variability, dataset characteristics, and model reliability are systematically reported. It can be extended to various EEG classification tasks, clinical research, and neuroscience applications. Through the introduction of this framework, we aim to establish a standard for transparent and reproducible model reporting, ultimately contributing to responsible AI adoption in EEG research and clinical settings.


 

\vspace{2em}
\section*{Acknowledgements}

The authors thank Harry Hallock and Aleksandar Babic (DNV, Norway) for useful discussions and suggestions around the topic of model cards and data cards. This work has received funding from the European Union's Horizon 2020 research and innovation program under grant agreement No. 964220. We conducted experiments on the Experimental Infrastructure for Exploration of Exascale Computing (eX3) system, financially supported by RCN under contract 270053.

\bibliographystyle{elsarticle-num}


\end{document}